\documentclass{article}

\usepackage{arxiv}

\usepackage[utf8]{inputenc} 
\usepackage[T1]{fontenc}    
\usepackage{hyperref}       
\usepackage{url}            
\usepackage{booktabs}       
\usepackage{amsfonts}       
\usepackage{nicefrac}       
\usepackage{microtype}      
\usepackage{lipsum}		
\usepackage{graphicx}
\usepackage{natbib}
\usepackage{doi}
\usepackage{float}

\graphicspath{ {./images/} }

\title{What Does DALL-E 2 Know About Radiology?}

\author{ 
    \href{https://orcid.org/0000-0001-5836-4542}{\includegraphics[scale=0.06]{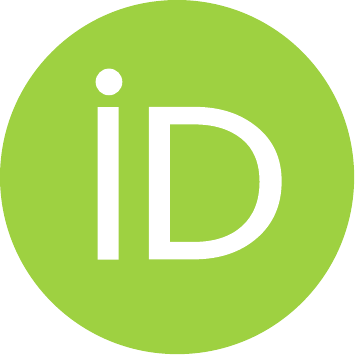}\hspace{1mm}Lisa C.~Adams}\thanks{contributed equally}\\
	Department of Radiology \\
	Stanford University School of Medicine \\
	Stanford, CA, USA \\
	\texttt{lcadams@stanford.edu} \\
	\And
	\href{https://orcid.org/0000-0001-9770-8555}{\includegraphics[scale=0.06]{orcid.pdf}\hspace{1mm}Felix~Busch}\footnotemark[1]\\
	Department of Radiology \\
	Charité – Universitätsmedizin Berlin \\
	Berlin, Germany \\
	\texttt{felix.busch@charite.de} \\
	\And	
	\href{https://orcid.org/0000-0002-9605-0728}{\includegraphics[scale=0.06]{orcid.pdf}\hspace{1mm}Daniel~Truhn}\\
	Department of Diagnostic and Interventional Radiology \\
	University Hospital Aachen \\
	Aachen, Germany \\
	\texttt{dtruhn@ukaachen.de} \\
	\And	
	\href{https://orcid.org/0000-0001-7719-8236}{\includegraphics[scale=0.06]{orcid.pdf}\hspace{1mm}Marcus R.~Makowski}\\
	Department of Diagnostic and Interventional Radiology \\
	School of Medicine and Klinikum Rechts der Isar \\
	Technical University of Munich \\
	Munich, Germany \\
	\texttt{marcus.makowski@tum.de} \\
	\And	
	\href{https://orcid.org/0000-0002-2122-2003}{\includegraphics[scale=0.06]{orcid.pdf}\hspace{1mm}Hugo JWL.~Aerts}\\
	Artificial Intelligence in Medicine (AIM) Program \\
	Mass General Brigham, Harvard Medical School \\
	Boston, MA, USA \\\\
    Departments of Radiation Oncology and Radiology \\
    Dana-Farber Cancer Institute and Brigham and Women's Hospital \\
    Boston, MA, USA \\\\
    Radiology and Nuclear Medicine, \\ 
    CARIM \& GROW, Maastricht University \\
    Maastricht, the Netherlands \\
	\texttt{HAerts@bwh.harvard.edu} \\	
	\And
	\href{https://orcid.org/0000-0001-9249-8624}{\includegraphics[scale=0.06]{orcid.pdf}\hspace{1mm}Keno K.~Bressem}\\
	Department of Radiology \\
	Charité – Universitätsmedizin Berlin \\
	Berlin, Germany \\\\
	Berlin Institute of Health at Charité – Universitätsmedizin Berlin \\ 
	Berlin, Germany \\
	\texttt{keno-kyrill.bressem@charite.de} \\\\
}
\date{}

\renewcommand{\headeright}{}
\renewcommand{\shorttitle}{What Does DALL-E 2 Know About Radiology?}

\hypersetup{
pdftitle={What Does DALL-E 2 Know About Radiology?},
pdfsubject={q-bio.NC, q-bio.QM},
pdfauthor={Keno K.~Bressem},
pdfkeywords={Deep Learning, DALL-E 2, Radiology, Diffusion Models},
}
\begin{document}
\maketitle
\begin{abstract}
Generative models such as DALL-E 2 could represent a promising future tool for image generation, augmentation, and manipulation for artificial intelligence research in radiology provided that these models have sufficient medical domain knowledge. Here we show that DALL-E 2 has learned relevant representations of X-ray images with promising capabilities in terms of zero-shot text-to-image generation of new images, continuation of an image beyond its original boundaries, or removal of elements, while pathology generation or CT, MRI, and ultrasound images are still limited. The use of generative models for augmenting and generating radiological data thus seems feasible, even if further fine-tuning and adaptation of these models to the respective domain is required beforehand.
\end{abstract}


\section{The potential impact of generative models in radiology}
DALL-E 2 is a novel deep learning model for text-to-image generation, first introduced by OpenAI in April 2022 \citep{ramesh2022}. The model has recently gained widespread public interest due to its ability to create photorealistic images solely from short written inputs \citep{kather2022}\citep{conwell2022}\citep{marcus2022}. Trained on billions of text-image pairs extracted from the internet, DALL-E 2 learned a wide range of representations, which it can recombine to create novel images, exceeding the variability it has seen in the training data, even from implausible prompts (e.g., “a painting of a pelvic x-ray by Claude Monet”)\citep{ramesh2022}\citep{nichol2021}.  
Given these powerful generative capabilities of DALL-E 2 it raises the question if these can be transferred to the medical domain to create or augment data, as it is often sparse and hard to come by. Given its size, it is to be assumed, that the training data for DALL-E 2 also contained radiological images and the model might have learned about composition and structure of X-rays and maybe even cross-sectional imaging and ultrasound. \\
Therefore, this comment systematically explores the radiological domain knowledge embedded in DALL-E 2, by creating and manipulating radiological images of X-rays, CT, MRI and ultrasound. 

\section{Generating radiological images from short descriptive texts}
To explore the general radiological domain knowledge embedded in DALL-E 2, we first tested the how well radiological images can be created from short text descriptive texts using the wording “An x-ray of” and one-word description of the anatomical area. For the head, chest, shoulder, abdomen, pelvis, hand, knee, and ankle, we let DALL-E 2 synthesize four artificial X-ray images, which are displayed in Figure 1. The coloring of the images and the general structure of the bones appeared realistic, and the overall anatomical propositions were correct, indicating the presence fundamental concepts of X-ray anatomy. However, upon close inspection, it was noted that the trabecular structure of the bone appeared random and did not follow the course of mechanical stress as in real X-rays. Sometimes smaller bones were either missing, like the fibula in multiple X-rays of the knee, or fused into a single instance like several carpal or tarsal bones. Still, the overall structure of the depicted anatomic area was recognizable. In rare cases, additional bones and joints were generated, e.g., an extra index finger in a hand X-ray. The model had the most difficulties generating joints correctly as they appeared to be either fused or the joint surfaces were distorted. \\
Compared to X-rays, the generation of cross-sectional images from CT or MRI was very limited (See Figure 2). While the requested anatomical structure was conceivable in the images, the overall impression was primarily nonsensical. Instead of a cross-sectional slice, a photo of an entire CT or MRI examination printed on radiographic film was often displayed. For ultrasound, the requested anatomical structures were not recognizable, instead all images resembled obstetric images. Still, the images showed concepts of CT/MRI/ultrasound images, indicating that representations of these modalities are present in DALL-E 2. 

\section{Changing areas in radiological images while maintaining the anatomical structure}
To investigate how detailed the radiological knowledge is, we tested how well DALL-E 2 can reconstruct missing areas in a radiological image (inpainting). Therefore, we chose radiographs of the pelvis, ankle, chest, shoulder, knee, wrist, and thoracic spine and erased specific areas before providing the residuals to DALL-E 2 for restoration (see Figure 3). The accompanying prompts were identical to the text-based imaging described above. DALL-E 2 provided realistic image replacements that were nearly indistinguishable from the original radiographs for the pelvis, thorax, and thoracic spine. However, the results were not as convincing when a joint was included in the image area. For the ankle and wrist, the number of tarsal bones and the structure varied greatly and deviated from realistic representations. For the shoulder, DALL-E 2 failed to reconstruct the glenoid cavity and articular surface of the humerus. In one case, a foreign body was incorporated into the shoulder that remotely resembled a prosthesis. For knee reconstruction, the model omitted the patella but retained the bicondylar structure of the femur.

\section{Extending radiological images beyond image borders}
As the previous two experiments showed, that DALL-E 2 possesses fundamentals of standard radiological images, we now aimed to explore how well anatomical knowledge is embedded in the model. For this we randomly selected X-rays of the abdomen, chest, pelvis, knee, different spine areas, wrist bones, and hand let DALL-E 2 extend these images beyond their borders, as this task requires the model to understand the location as well as anatomical proportions. The accompanying prompt was chosen depending on the anatomical regions being generated. Again, the style of the augmented images represented a realistic depiction of X-ray (see Figure 4). It was possible to generate a complete whole-body X-ray using just an image of the knee as a starting point. However, the further the distance between the original image and the generated area, the less detailed the images became. Anatomical proportions, such as the length of the femoral bone or size of the lungs, remained realistic, still finer details such as the number of lumbar vertebrates were inconsistent. The model generally performed best when creating anterior and posterior views, while lateral views were more challenging and produced poorer results. 

\section{Generation of pathological images}
The generation of pathological images, e.g., showcasing fractures, intracranial bleedings, or tumors was limited. We tested this with the generation of bone fractures on X-ray, which mostly led to distorted images of normal radiographs, similar to those shown in Figure 1. Furthermore, DALL-E 2 has a filter that prevents the generation of harmful content, preventing us from testing most pathologies, as words like “hemorrhage” triggered this filter. 

\section{Conclusion and Outlook}
We could show that DALL-E 2 can generate X-ray images that are close to authentic X-ray images in style and anatomical proportions. From this, we conclude that relevant representations for X-rays were learned by DALL-E 2 during training. However, it seems only representations for radiographs without pathologies are present or admitted by the filter, as generating pathological images, e.g., of fractures and tumors, was limited. Beyond, generative capabilities for cross-sectional imaging as well as ultrasound were poor. \\
Access to data is critical in deep learning and generally, the larger the dataset, the better the performance of the models trained upon it. However, especially in radiology, there exists not unified large database from which such dataset could be created but instead data is split between multiple institutions and privacy concerns additionally prevent merging of this data. Here, synthetic data brings the promise to solve these issues, allowing the creation much larger datasets than those currently available substantially accelerating the development of new radiological deep learning tools \citep{jordon2020}\citep{jordon2022}. 
Previous approaches on data generation using generative adversarial networks (GANs) were challenged by instabilities during training and the inability to capture sufficient diverytiy of the training data \citep{kodali2017}. In contrast novel diffusion based generative models, of which DALL-E 2 is a prominent representative, solved these issues, beating GANs on image generation \citep{dhariwal2021}. Furthermore, the amount of control and flexibility diffusion models can provide during data generation, makes them exciting for radiological data augmentation, as images may be altered, and pathologies may be added or removed just by text prompts. However, our experiments showed, that DALL-E 2 was not yet able to create convincing pathological images, which is likely explained by a lack of those images in the training data. Radiological images containing a wide variety of pathologies are more likely to be shared on professional sites, might be under restricted access or in special data formats such as DICOM. Altogether probably leading to only a small number of those images included on the training data. \\
Therefore, more work is needed to explore diffusion models designed to create and modify medical data. Although DALL-E 2 is not openly available, other models of similar architecture, such as Stable Diffusion, are available to the public\citep{ho2022}\citep{saharia2022}. Future research should focus on fine-tuning these models to medical data and incorporate medical terminology to create powerful models for data generation and augmentation in radiological research. 

\newpage
\bibliographystyle{unsrtnat}
\bibliography{references}  

\newpage
\section{Figures}
\begin{figure}[H]
	\centering
    \includegraphics{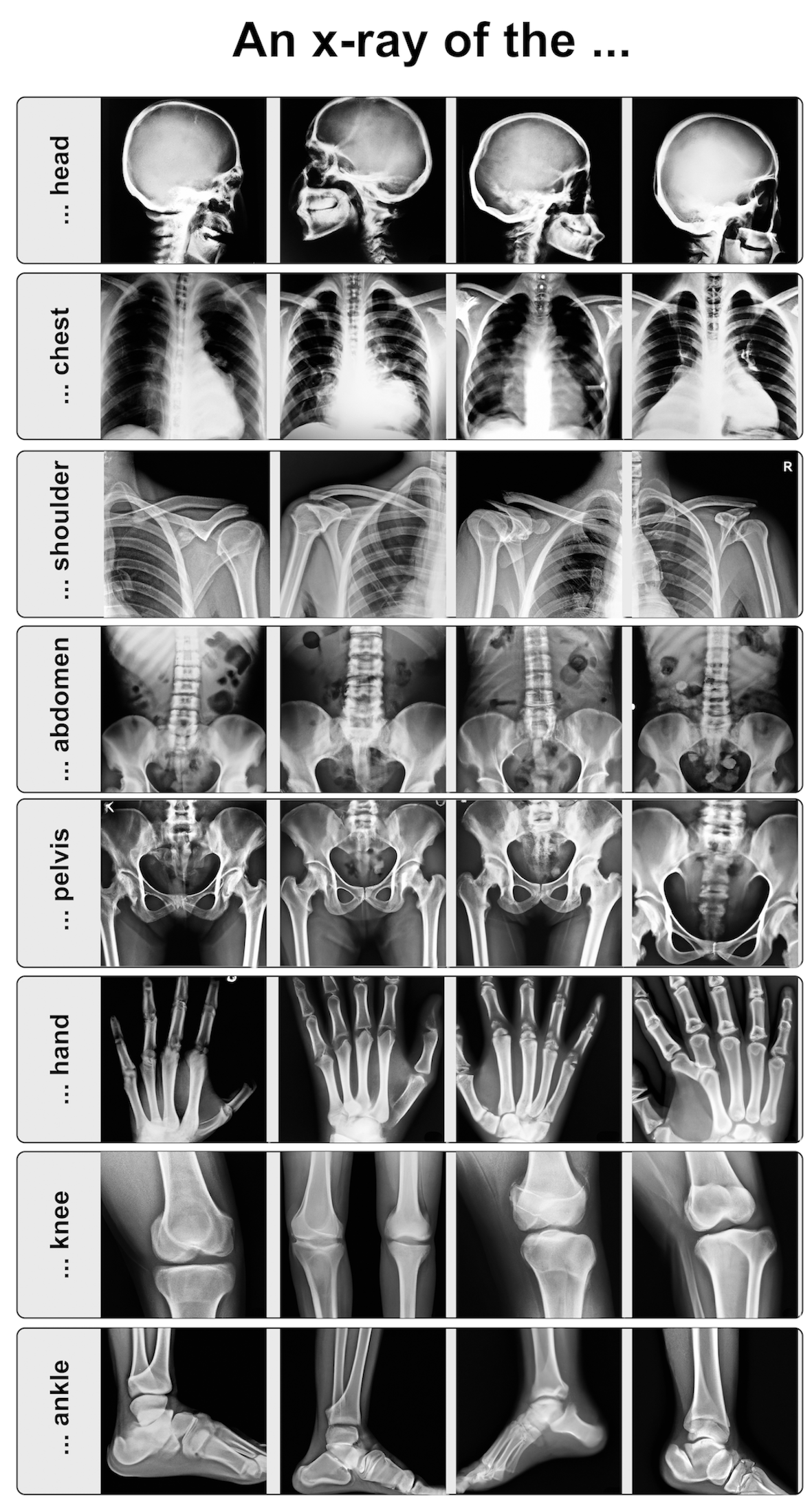}
	\caption{Sample generated anatomical structures in X-ray from short text descriptions using DALL-E 2.}
	\label{fig:fig1}
\end{figure}

\begin{figure}[H]
	\centering
    \includegraphics{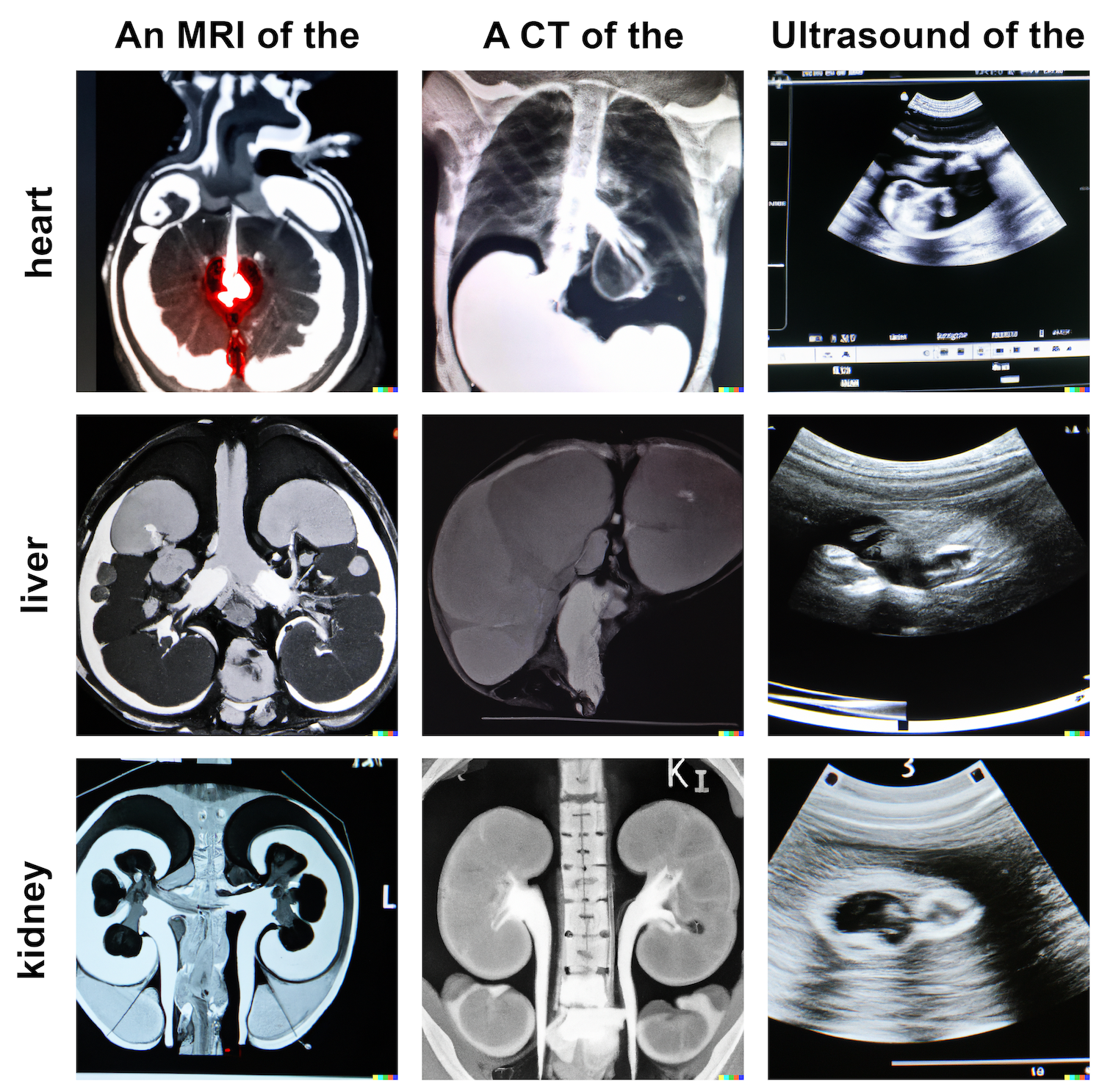}
	\caption{Sample text-to-image generated anatomical structures in CT, MRI, and ultrasound using DALL-E 2.}
	\label{fig:fig2}
\end{figure}

\begin{figure}[H]
	\centering
    \includegraphics{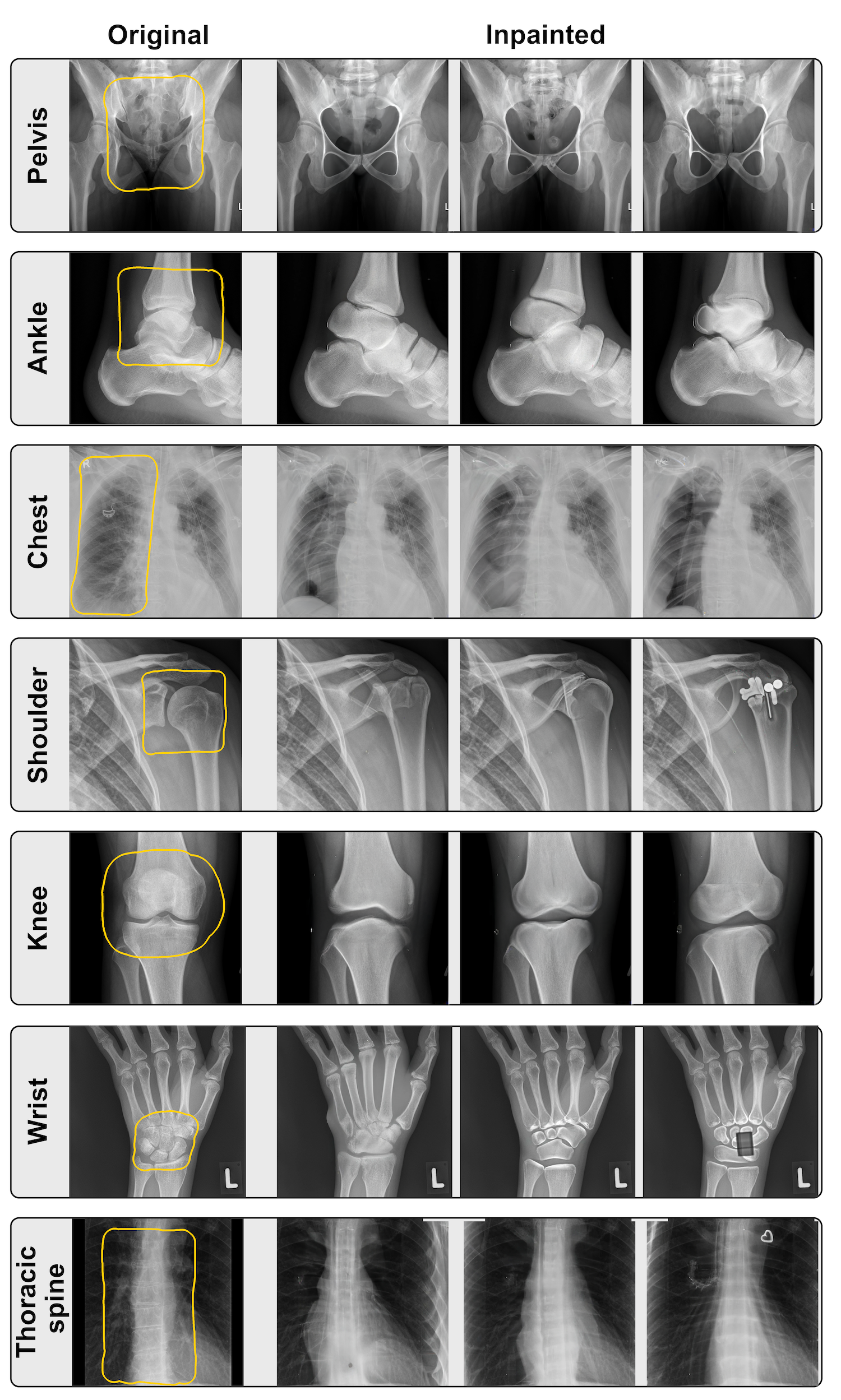}
	\caption{Reconstructed areas of different anatomical locations in X-rays using DALL-E 2. The yellow-bordered regions of the original images were erased before providing the remnant image for reconstruction.}
	\label{fig:fig3}
\end{figure}

\begin{figure}[H]
	\centering
    \includegraphics{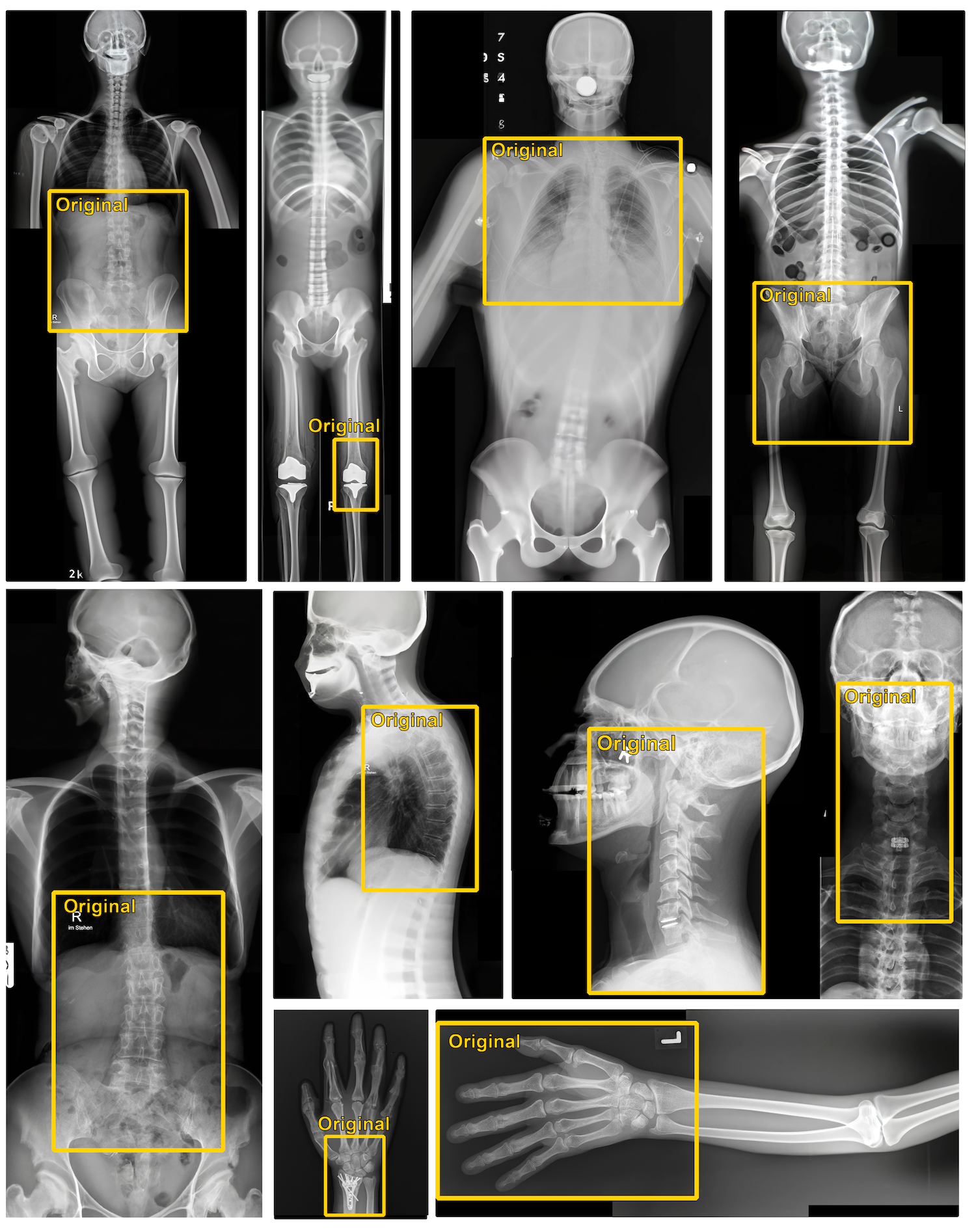}
	\caption{Extending X-ray images of different anatomical regions beyond their borders using DALL-E 2. The original X-ray is shown in the yellow boxes, with the remainder of the images being generated by DALL-E 2.}
	\label{fig:fig4}
\end{figure}

\end{document}